\documentclass{article}

\usepackage{arxiv}

\usepackage[utf8]{inputenc} 
\usepackage[T1]{fontenc}    
\usepackage{hyperref}       
\usepackage{url}            
\usepackage{booktabs}       
\usepackage{amsfonts}       
\usepackage{nicefrac}       
\usepackage{microtype}      
\usepackage{lipsum}
\usepackage{graphicx}
\graphicspath{ {./images/} }
\usepackage{amsmath}
\usepackage{tikz}
\usepackage{BOONDOX-uprscr}
\usepackage{amsthm}
\usepackage[numbers]{natbib}
\usepackage{subcaption}
\usepackage{array} 
\usepackage{booktabs}
\usepackage[vlined,ruled,linesnumbered]{algorithm2e}

\usepackage[utf8]{inputenc}
\usepackage[english]{babel}

\usepackage{natbib}

\title{A Bilevel Optimization Framework for Imbalanced Data Classification}

\author{
 Karen Medlin \\
  University of North Carolina at Chapel Hill \\
  \texttt{kmedlin@unc.edu} \\
   \And
 Sven Leyffer \\
  Argonne National Laboratory \\
  \And
 Krishnan Raghavan \\
  Argonne National Laboratory \\
}

\begin{document}
\maketitle
\begin{abstract}
Data rebalancing techniques, including oversampling and undersampling, are a common approach to addressing the challenges of imbalanced data. To tackle unresolved problems related to both oversampling and undersampling, we propose a new undersampling approach that: (i) avoids the pitfalls of noise and overlap caused by synthetic data and (ii) avoids the pitfall of under-fitting caused by random undersampling. Instead of undersampling majority data randomly, our method undersamples datapoints based on their ability to improve model loss. Using improved model loss as a proxy measurement for classification performance, our technique assesses a datapoint's impact on loss and rejects those unable to improve it. In so doing, our approach rejects majority datapoints redundant to datapoints already accepted and, thereby, finds an optimal subset of majority training data for classification. The accept/reject component of our algorithm is motivated by a bilevel optimization problem uniquely formulated  to identify the optimal training set we seek. Experimental results show our proposed technique with F1 scores up to 10\% higher than state-of-the-art methods. 
\end{abstract}

\section{Introduction}
Technological advances have ushered in an era when all the data generated for before, during, and after a scientific experiment is kept. Such vast quantities of stored data have changed the landscape of analysis where careful attention must be paid to new statistical challenges introduced by massive amounts of data~\cite{fan2014challenges}. For instance, a billion subatomic particle interactions take place over a period of 24 hours in the ATLAS particle detector. However, only about 2,500 of these interactions are useful and must be extracted through careful analysis. This problem of identifying a small amount of useful events from a large pool of interactions is often modelled as an imbalanced data classification problem~\cite{cao2021heteroskedastic}. 

\begin{figure}
\centering

\begin{subfigure}{0.7\textwidth}
\centering
\includegraphics[width=0.7\linewidth]{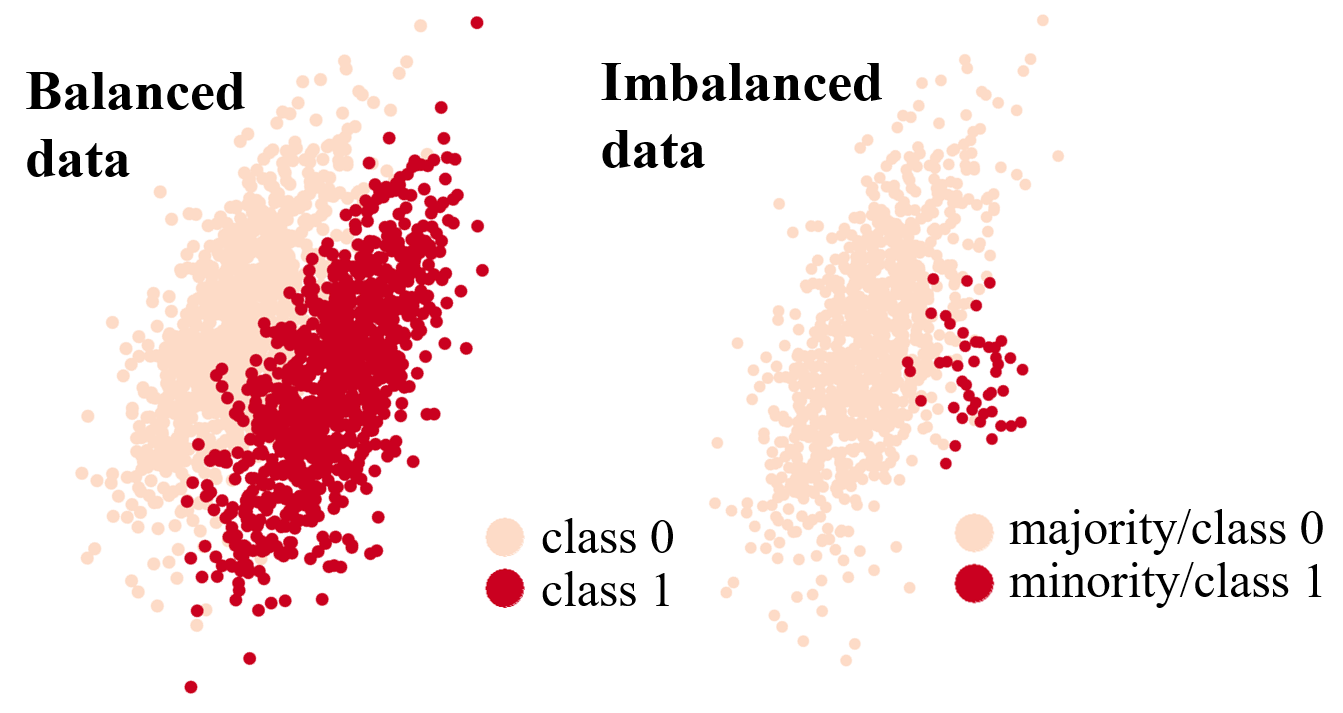}
        \caption{Without intervention, imbalanced data trains in favor of the majority class. However, it is the minority class - diagnostic medical images, undiscovered drug compounds, threats to network security and other under-represented yet critical scenarios - we want to know more about.}
        \label{fig:imbalancedDots}
\end{subfigure}
\vspace{2cm}
\\
\begin{subfigure}{0.7\textwidth}
\centering
\includegraphics[width=0.7\linewidth]{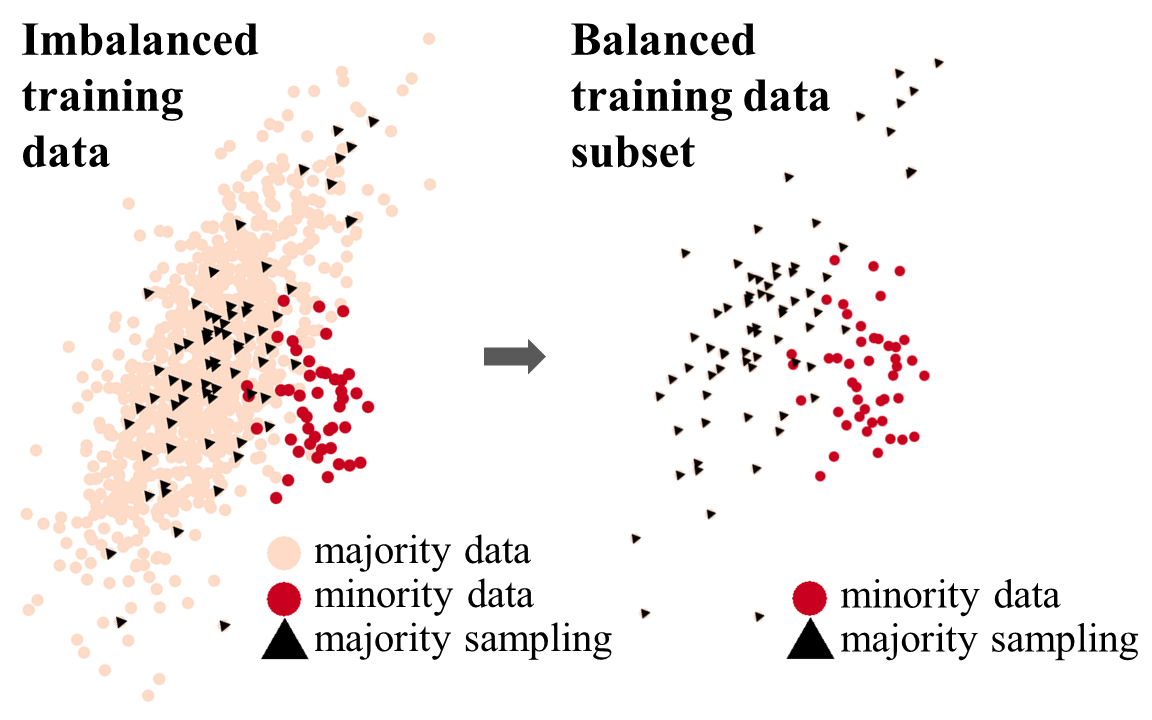}
        \caption{A bilevel optimization framework provides a mechanism for undersampling majority data to attain an optimal  subset of balanced training set. } 
        \label{fig:Mom}
\end{subfigure}
\end{figure}

A dataset is imbalanced when one of the categories represented in the data is strongly under-represented. As depicted in Figure \ref{fig:imbalancedDots}, the under-represented class is called minority data whereas the over-represented class is called majority data. Other examples of imbalanced data sets include diseases \cite{bria2020addressing,johnson2019survey, khan2018cost-sensitive}; chemicals either harmful or beneficial to humans \cite{bae2021effective, korkmaz2020chemicaljournal}; natural disasters \cite{kumar2021classification}; fraudulent transactions \cite{singh2022credit, makki2019experimental} and many more \cite{kanika2020survey, haixiang2017learning}. While it is the rare minority data that is of interest, the sparsity of the minority data within the larger dataset presents challenges to neural network models.

A fundamental problem of classification is the  impairment that imbalanced data causes to training. Without intervention and as depicted in Figure \ref{fig:Mom}, training sets of imbalanced data are also imbalanced. Without an equal amount of minority and majority data for training, a model is impaired in its ability to classify the one it lacks - the minority data.

For over 20 years, the landscape of imbalanced data classification has been dominated by the Synthetic Minority Oversampling Technique (SMOTE) \cite{chawla2002smote} and its derivatives. SMOTE and SMOTE-like methods tackle the problem by generating synthetic minority data. The synthetic data is used to rebalance the training data so that it contains equal amounts of majority and minority data. SMOTE is an example of oversampling. 
 
While oversampling methods address the problem of the lack of minority data, they fall short of solving the problem of imbalance. Using synthetic data to augment the minority population introduces noise and overlap. Synthetic data is noisy in that it does not entirely match real data. The additional minority datapoints, in the form of synthetic data, increases the possibility of overlap between minority and majority data. Both noise and overlap make it difficult for a model to distinguish the two types of data from one another.

To reduce the issues due to noise and overlap, a tool originally built to generate realistic images called Generative Adversarial Networks (GANs)\cite{goodfellowGenerative2014} has been repurposed. Specifically, GANs have been repurposed to leverage their adversarial structure to reject synthetically generated datapoints that do not help with classification  \cite{sharmaSmotifiedGan2022, cheahEnhancing2023, ahsanEnhancing2024}. More specifically, GANs reject synthetic datapoints that do not decrease classification loss. Despite improvements made with the use of GANs, problems of noise and overlap persist when using oversampling methods to classify imbalanced data.

While GANs help with eliminating synthetic datapoints that do not improve model loss, they do not necessarily eliminate synthetic datapoints contributing to noise and overlap. Given that the number of different features making up a single datapoint can be quite large, and that each individual feature can also have a large number of possible values, there is a vast sea of possibilities for any one datapoint. Therefore, creating synthetic data that matches minority data in a way that distinguishes it from majority data while also avoiding problems of over-fitting is a challenge difficult to overcome.   

Sidestepping the problems caused by synthetic data can be achieved by \textit{under}sampling majority data instead of \textit{over}sampling minority data. While both techniques \text{rebalance} the data, majority undersampling avoids problems brought on by synthetic data by simply not using any. Instead, they use fewer majority datapoints to balance with the dataset's lower number of minority datapoints. However, the common undersampling approach of random undersampling is marred by problems of under-fitting. Under-fitting occurs when training data does not reflect the distribution of the original dataset and, therefore, impairs training. When majority undersampling is arbitrary, the resulting majority training set is likely to miss datapoints critical to capturing the majority dataset's distribution. 
 
To address unresolved problems of both oversampling and undersampling techniques, we propose a new majority undersampling method that: (i) avoids the pitfalls of noise and overlap caused by synthetic data, and (ii) avoids the pitfall of under-fitting caused by random undersampling. Instead of undersampling majority data randomly, our approach rejects datapoints based on their inability to improve model loss similar to how GANs reject synthetic data. Using improved model loss as a proxy measurement for classification performance, GANs assess a datapoint's impact on classification performance and reject those unable to decrease model loss. Instead of rejecting synthetic datapoints, our proposed technique rejects majority datapoints. As such, it rejects majority datapoints redundant to majority datapoints already accepted and, thereby, unable to improve majority loss.

Instead of using a GAN for the rejection step, our approach uses a bilevel optimization framework. The lower level problem of our bilevel problem minimizes model loss over model parameters. The upper level problem uniquely minimizes majority loss over subsets of majority data. Motivated to solve the upper level problem, our approach randomly selects subsets of majority data to be added to the training set, and then either accepts or rejects them based on the subset's ability or inability to improve majority loss. With its collection of accepted subsets of majority data, our method formulates an ideal set of training data that improves imbalanced data classification. While bilevel optimization has been used in the context of imbalanced data classification, to the best of our knowledge it has not been used to rebalance training data. 

By not using synthetic data, our method improves upon unresolved problems of noise and overlap of oversampling methods. By using a bilevel optimization framework to select an optimal subset of majority training data, our approach improves upon under-fitting problems common to undersampling methods.

Our contributions can be summarized as follows:
\begin{itemize}
\item A novel approach to imbalanced data classification using bilevel optimization. The bilevel optimization framework allows for both optimizing model parameters and indentifying an optimal subset of majority training data.
\item A novel algorithm for classifying imbalanced data.  Called MUBO: Majority Undersampling with Bilevel Optimization, experimental results are competitive with state-of-the-art sampling methods. 
\item An analysis that shows an increased number of correctly classified majority datapoints implies improved majority loss. This result supports MUBO's undersampling approach that rejects majority datapoints based on their inability to decrease majority loss.
\end{itemize}

\subsection{Related Works}
\label{relatedWorks}
Approaches to imbalanced data classification generally fall into one of two  groups: those that rebalance the data used to train the model - as SMOTE does - and those that rebalance an aspect of the neural network model itself. Common methods that rebalance the model include those that reweight the model's loss function to make up for its bias towards majority data \cite{dangInverse2024, pmlrV162Guo22e}. In what follows, we focus our review on methods that rebalance training data given that our method takes this approach. More exhaustive reviews of imbalanced data classification methods can be found in recent surveys \cite{strelceniaSurvey2023, dasSupervised2022, kovacsEmpirical2019}.

As previously mentioned, SMOTE and its derivatives, including SMOTE+TomekLinks \cite{batistaStudy2004}, Borderline-SMOTE \cite{hutchisonBorderlineSmote2005} and SVM-SMOTE \cite{nguyenBorderline2011}, have dominated the landscape of imbalanced data classification for over 20 years. Despite their impressive performance, their approach of generating synthetic data has yet to resolve the challenge of imbalanced data classification. Persistent problems common to using synthetic data include noise and overlap. These problems are addressed with more recent methods such as Reduced-Noise SMOTE \cite{badawyRnSmote2022} that removes noise after oversampling and Radius-SMOTE \cite{PradiptaRadiusSMOTE} that prevents overlap among generated samples. A lack of diversity of minority data is another unresolved issue encountered with oversampling approaches. A lack of diversity of the synthetic data causes the model to over-fit to the training data and, in turn, impairs classification. Sampling methods that address issues of over-fitting include Diversity-based Selection \cite{yangDiversityBased2023} that uses matrix inverses to measure how much diversity a datapoint contributes and BalanceMix \cite{songToward2023} that interpolates minority class datapoints with datapoints that have been augmented using Mixup \cite{zhangMixup2018}. Additionally, RIO \cite{changImageLevel2021} increases the diversity of training sets by labeling both images and common objects in the images called ``sub-images.''

Also previously mentioned, a second overarching tool being used to address the problem of imbalance are GANs. Specifically, GANs are being used to address problems of noise, overlap and over-fitting that arise due to SMOTE-based oversampling techniques. For example, CTGAN-MOS \cite{MajeedCTgan} features steps to reduce noise and improve the quality of training data. In fact, several recent approaches use both SMOTE and GANs. The SMOTE component generates synthetic datapoints, and the GAN rejects those synthetic datapoints that do not improve model loss. For example, SMOTified-GAN \cite{sharmaSmotifiedGan2022} employs SMOTE-generated samples as input for a GAN instead of random numbers. GAN-based Oversampling and SVM-SMOTE-GAN build on SMOTified-GAN's approach to address issues of noise, overlap and over-fitting common to oversampling techniques. To decrease noise and increase diversity of synthetic data, GAN-based Oversampling (GBO) \cite{ahsanEnhancing2024} trains a generator network to replicate the distribution of the original data. SVM-SMOTE-GAN (SSG) \cite{ahsanEnhancing2024} interpolates between existing minority class samples to generate synthetic data. 

Despite efforts to overcome problems of noise, overlap and over-fitting, these challenges persist with oversampling methods such as SMOTified-GAN, GBO and SSG. There is a limit on how well these methods are able to classify minority data due issues associated with synthetic data. In contrast, our proposed approach addresses unresolved oversampling issues by not relying on synthetic data. Instead, our method holds minority data constant while undersampling majority data. Additionally, it addresses common problems associated with undersampling such as under-fitting by optimizing over subsets of majority training data using bilevel optimization. Experimental results in section \ref{results} compare our approach with SMOTified-GAN, GBO and SSG.  

To the best of our knowledge, while bilevel optimization is not new in the context of imbalanced data classification \cite{rosalesPerezHandling2023, chabbouhImbalanced2023, hammamiFeature2020}, there is no work in the literature that uses bilevel optimization to rebalance training data. On the other hand, established methods are similar to the formulation of the lower level of our bilevel problem that minimizes loss over model parameters. For example, AutoBalance \cite{liAutobalance2022} classifies imbalanced data with a bilevel optimization framework that minimizes model loss over parameters and hyperparameters. Additionally, network architecture search methods formulate their bilevel optimization problems with lower level problems that minimize model loss \cite{zhouTheoryInspired2020, liuDarts2019, luketinaScalable2016}. Our unique use of bilevel optimization resides in the formulation of our upper level problem as depicted in Figure \ref{fig:bilevelFig} and further discussed in Section \ref{bilevelApproach}.

\section{Background}
\label{background}
This section provides background for our bilevel optimization problem and the motivation behind it in Sections \ref{bilevelApproach} and \ref{algo} and for evaluation metrics and analysis in Section \ref{evaluation}. 

We begin with a set of imbalanced data $\mathscr{S} \subset \mathbb{R}^n \times \mathbb{R}^1$. As the goal is to classify $\mathscr{S}$ into majority and minority subsets, let $\mathscr{S} = \mathscr{S}_m \cup \mathscr{S}_M$ where $\mathscr{S}_m\subset \mathbb{R}^n \times \mathbb{R}^1$ is the minority data, and $\mathscr{S}_M\subset \mathbb{R}^n \times \mathbb{R}^1$ is the majority data. To train a model, let $S=S_m \cup S_M$ be a sample of training data where $S_m\subset \mathscr{S}_m$ is the minority subsample, and $S_M\subset \mathscr{S}_M$ is the majority subsample. Let $m=|S_m|$ and $M=|S_M|$ denote the number of minority and majority datapoints; $m, M \in \mathbb{N}$. As $\mathscr{S}$ is imbalanced, assume $S$ is imbalanced and $m<<M$.

To track classification and measure model loss and performance, let a datapoint of the training sample be denoted with $s=(x,y)\in S$ where the feature component is denoted with $x\in \mathbb{R}^n$, and the label component is denoted with $y \in \{0,1\}$. As is customary, $y=0$ denotes the majority or `negative' class while $y=1$ denotes the minority or `positive' class. Without loss of generality (w.l.o.g.), let $s_1,\dots,s_m \in S_m$ and $s_{m+1}, \dots, s_{m+M} \in S_M$. 

To predict the label of an unlabeled $x$, we use a multinomial logistic classifier called \textit{softmax}. Let $z(w) \in \mathbb{R}$ be the output of a convoluted neural network with weights $w$ dependent on a sample of training data $S$.
 
\newtheorem{definition}{Definition}
\begin{definition}\label{def:softmax}
Let $\text{softmax}(s;z(w)) \in \mathbb{R}^{1 \times 2}$ denote \textbf{majority and minority classification probabilities} for a datapoint $s \in S$ as follows

\begin{footnotesize}
\begin{align*}
    \text{softmax}(s;z(w)) &= \big[p(y = 0 |x), \hspace{.1in} p(y = 1 |x)\big]\\
    &= \left[\frac{e^{1-z(w)}}{ e^{z(w)} + e^{1-z(w)}},\hspace{.1in}\frac{e^{z(w)}}{ e^{z(w)} + e^{1-z(w)}}\right].\\
\end{align*}
\end{footnotesize}
\end{definition}
Classification depends on which of the two classes $s$ is most likely to belong; i.e., which classification probability is larger. With $\hat{y} \in \{0,1\}$ being the predicted class,
\begin{footnotesize}
\begin{align*}
    \hat{y} &= \underset{j \in \{0,1\}}{\text{argmax}} \quad   p(y = j |x)\\
    &= \begin{cases} 
        0 \quad \text{if} \quad e^{1-z(w)} \geq e^{z(w)}\\ 
        1 \quad \text{if} \quad e^{1-z(w)} < e^{z(w)}.
    \end{cases}\\
\end{align*}   
\end{footnotesize}

While classification depends on which of the two classes $s$ most likely belongs, loss depends on which of the two classes $s$ actually belongs. Defined below, MUBO's loss function is commonly referred to as ``binary cross entropy'' \cite{bishop1996}.
\begin{definition} \label{def:loss}
    Let $l(s;z(w)) \in \mathbb{R}^+$ denote \textbf{loss} as follows
\begin{footnotesize}
\begin{align*}
    l(s;z(w)) 
    &= \begin{cases} 
        -\log(\frac{e^{1-z(w)}}{ e^{z(w)} + e^{1-z(w)}}), \quad \text{if  } s \in S_M; \text{ i.e.}, y=0\\ 
        -\log(\frac{e^{z(w)}}{ e^{z(w)} + e^{1-z(w)}}), \quad \text{if  } s \in S_m; \text{ i.e.}, y=1\\
    \end{cases}
\end{align*}    
\end{footnotesize}
\end{definition}

\begin{definition}
Let $\mathscr{L}(S; z(w))  $ denote \textbf{sample loss} as follows 
\begin{footnotesize} 
\begin{align*}
    \mathscr{L}(S; z(w)) &= \frac{1}{|S|}\sum_{i=1}^{|S|}l(s_i; z(w))\\ 
    &= \underbrace{\frac{1}{|S|}\sum_{i \in S_m}l(s_i; z(w))}_{J(S_m; z(w))}\quad + \underbrace{\frac{1}{|S|}\sum_{i \in S_M}l(s_i; z(w))}_{J(S_M; z(w))},\\
\end{align*}    
\end{footnotesize}
where $J(S_m; z(w))$ and $J(S_M; z(w))$ denote \textbf{minority loss} and \textbf{majority loss}, respectively. 
\end{definition}

To measure a classifier's performance, let $T_p \in \mathbb{N}$ and $F_n\in \mathbb{N}$ denote the number of correctly (truly) and incorrectly (falsely) classified minority datapoints, respectively; which implies $m=T_p + F_n$. Similarly, let $T_n\in \mathbb{N}$ and $F_p\in \mathbb{N}$ denote the number of correctly (truly) and incorrectly (falsely) classified majority datapoints; $M = T_n + F_p$. w.l.o.g., let $\hat{y}_1,\dots,\hat{y}_m$ be the predicted labels of minority datapoints and $\hat{y}_{m+1},\dots,\hat{y}_{m+M}$ be the predicted labels of majority datapoints. Therefore, 
$T_p = \sum_{i=1}^m \hat{y}_i$, $F_n = m - T_p$, $F_p = \sum_{i=m+1}^{m+M} \hat{y}_i$ and $T_n = M - F_p$.

Based on amounts of truly ($T_n/T_p$) and falsely ($F_n/F_p$) classified data, the F1 score is commonly used to measure model performance. Unlike accuracy, the F1 score is considered to be an unbiased and, therefore, preferred metric for imbalanced data classification.

\begin{definition}\label{def:F1}
Let $\text{F1} \in \mathbb{R}^+$ denote the \textbf{average F1 score} 

\begin{footnotesize}
\begin{align*}
    \text{F1} &= \frac{1}{2}\left(\underbrace{\frac{2T_p}{2T_p + F_p + F_n}}_{\text{F1}_m} + \underbrace{\frac{2T_n}{2T_n + F_n + F_p}}_{\text{F1}_M}\right),\\
\end{align*}
\end{footnotesize}
where $\text{F1}_m$ and $\text{F1}_M$ denote the \textbf{minority F1 score} and \textbf{majority F1 score}, respectively.\\
\end{definition}
We will use $\mathscr{L}(S; z(w))$ and $J(S_M; z(w))$ in the formulation of our bilevel optimization problem described in Section \ref{bilevelApproach}. To discuss majority loss and provide motivation for our bilevel optimization problem in Section \ref{algo}, we distinguish between the loss of truly classified majority points, $s^T \in S_M$, and of those falsely classified, $s^F \in S_M$, below. 
\begin{definition} \label{def:tfMajLoss}
Let $J^T_M$ denote \textbf{loss of truly classified majority} and $J^F_M$ denote \textbf{loss of falsely classified majority} as follows
\begin{footnotesize} 
\begin{align*}
    J^T_M &= \frac{1}{T_n}\sum_{s^T \in S_M}l(s^T; z(w))
\end{align*}
\begin{align*}
    J^F_M &= \frac{1}{F_p}\sum_{s^F \in S_m}l(s^F; z(w)),
\end{align*}
\end{footnotesize} 
\end{definition}
where $T_n$ and $F_p$ denote the number of truly and falsely classified majority datapoints as previously noted.

\section{Bilevel Optimization Formulation}
\label{bilevel}
\begin{figure}
\centering
\includegraphics[width=1.0\textwidth]{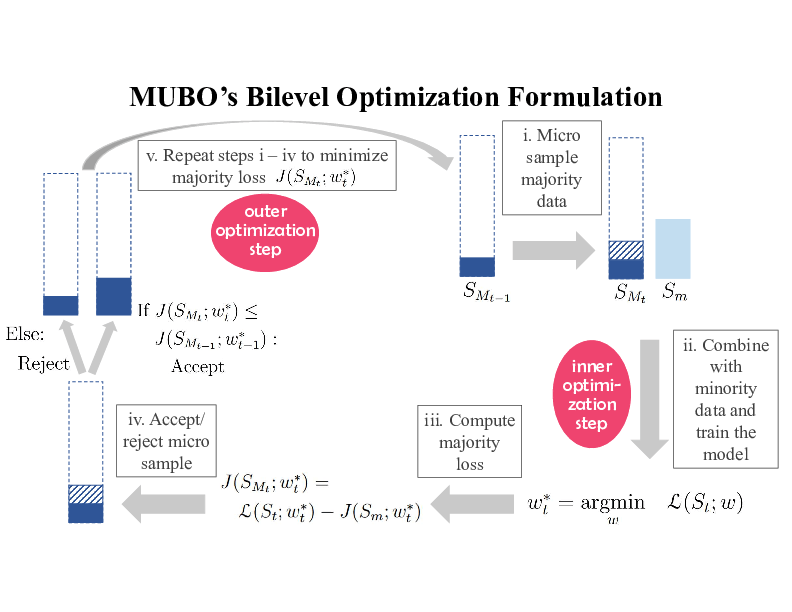}
    \caption{MUBO is designed to solve its bilevel optimization problem. One full iteration is shown above with the following steps: i. micro sample majority training data (the number of datapoints in the micro-sample is significantly less than the number of minority datapoints); ii. train the model using the new majority micro-sample combined with the previously accepted micro-samples of majority data and all minority training data, $S_t = S_{M_t} \cup S_m$, finding optimal weights, $w^*_t$,  that minimize overall loss (\textbf{inner optimization step}); iii. compute majority loss using the optimal weights, $J(S_{M_t}; w_t^*) = \mathscr{L}(S_t; w_t^*) - J(S_m;w_t^*)$; iv. compare majority loss of the current step $t$ with that of step $t-1$, accepting the new micro-sample of majority training data \textit{only if} it improves majority loss, i.e., $J(S_{M_t}; w_t^*)\leq J(S_{M_{t-1}};w_{t-1}^*)$; and v. repeat steps i-iv to find a subset of majority training data that minimizes majority loss (\textbf{outer optimization step}).}
    \label{fig:bilevelFig}
\end{figure}

We formulate the bilevel optimization problem at the heart of the MUBO algorithm to find an optimal subset of training data as follows
\begin{center}
    $\underset{S_M \subset \mathscr{S}}{\text{min}} \quad   J(S_M;z(w^*))$\\
    $\text{ subject to } w^* = \underset{w}{\text{argmin}} \quad \mathscr{L}(S;z(w))$\\
\end{center}
where $S = S_m \cup S_M$. Figure \ref{fig:bilevelFig} depicts the connection between MUBO and its bilevel problem.

The outer level problem chooses an optimal majority subsample, $S_M$, to minimize the majority loss function, subject to finding optimal neural network weights to minimize the sample loss $\mathscr{L}(S;z(w))$. Bilevel optimization is also referred to as leader-follower games and is a sequential decision model where the follower (inner level problem) chooses optimal weights given the leader's (outer level) choice of sample set. In this sense, the leader can anticipate the follower's decision and takes the optimal weights into account when choosing the sample set $S_M$. 

Our unique use of bilevel optimization resides in the formulation of the outer level problem. By focusing the outer level problem solely on finding an optimal subset of majority training data, we increase the likelihood of correctly classifying majority data. Additionally, using optimal majority training data, in turn, also increases the likelihood of correctly classifying minority data as there are only two classes. In turn, the opportunity for overlap between majority and minority data decreases. Attaining a noise-less and balanced subset of training data, we pair the resulting optimal subset of majority data with the full set of minority training data.  

To note, a bilevel optimization framework lends itself well to other classification methods and learning more generally. The outer level problem, in particular, can be swapped out for achieving other learning objectives.

\section{MUBO: Majority Undersampling with Bilevel Optimization}
\label{algo}
\label{bilevelApproach}

Reflective of the bilevel optimization problem it solves, MUBO features two loops as described in Algorithm \ref{MUBO}. Focused on the lower level problem, the inner loop trains the model and generates the parameters $w^*$ associated with minimized sample loss; i.e., $\underset{w}{\text{argmin}} \quad \mathscr{L}(S;z(w))$. 

Focused on the upper level problem, the outer loop randomly selects subsamples of the majority training data that are then added to a training set comprised of all the minority training data and previously accepted subsamples of majority data. This combined training set is used to train the model in the inner loop. The outer loop continues with its accept/ reject step that accepts only those subsamples of majority data able to improve majority loss. As the algorithm iterates, the combined training set, $S = S_m \cup \Delta S_M$, grows with the addition of accepted subsamples of majority data. The ultimate output is the optimal combined sample made up of all the minority training data and the optimal undersampling of majority training data that has not increased majority loss.

\begin{algorithm}[!htb]
\DontPrintSemicolon
\KwIn{Training data $S=S_m \cup S_M$}
\KwOut{Optimal training subset $S_t$}
\SetKwBlock{Begin}{function}{end function}

$S_{-1} = S_m$, $J(S_{M_{-1}}) = 10$, $M_0 = \lceil\frac{1}{10}m\rceil$, $t=0$\;

\SetKwBlock{Begin}{function}{end function}

\While{$t<$ MaxIter}
{
    $S_{M_t} =$ $S_M[\text{randint}(M_t)]$\;
    $S_t = S_{t-1} \cup S_{M_t}$\;
    $J(S_{M_t}) = \text{mean.innerLoop(}S_t) - \text{ mean.innerLoop(}S_t)[1 : m]$\;
    \If{(\text{checkStep}${(}J{(}S_{M_t}{)},J{(}S_{M_{t-1}}{)}{)})=$ reject}
    {
        $S_t = S_{t-1}$\;
        $J(S_{M_t}) = J(S_{M_{t-1}})$\;
    }
    \If{$|S_{t-1}| \leq \frac{7}{4}m$} {$M_{t+1} = \lceil\frac{1}{10}m\rceil$\;
        \Else{$M_{t+1} = \lceil\frac{1}{100}m\rceil$}}
$t = t + 1$\;
\Return $S_t$
}
\Begin($\text{innerLoop}{(}S_t{)}$)
{\While {$\lVert\nabla \mathscr{L}\rVert >$ \text{GradTol and epochs } $<$ \text{MaxEpochs}}{
    Optimize  $\mathscr{L}=\frac{1}{m+M_t}\sum_{s\in S}l(s)$\;
    $epochs = epochs + 1$\;
}
\Return{$\mathbf{l(s)}[1:m+M_t]$} 
}
\Begin($\text{checkStep}{(}J{(}S_{M_t}{)},J{(}S_{M_{t-1}}{)}{)}$)
{
    \If{$J(S_{M_t}) > J(S_{M_{t-1}})$}{\Return reject}
}

\caption{MUBO with its outer loop for sampling and then accepting/rejecting majority training data and inner loop for optimizing model parameters}\label{MUBO}
\end{algorithm}

\subsection{Motivation for the Upper Level Problem}
As minimizing loss over model parameters is common, motivation for our lower level optimization problem is clear. In this section, we provide theoretical results supporting the formulation of our upper level problem. We show that using only subsamples of majority data that decrease majority loss is a reasonable approach to improving classification. This result follows from the observation that the loss of correctly classified majority datapoints, $J_M^T$, is strictly less than the loss of incorrectly classified majority datapoints, $J_M^F$; as defined in Definition \ref{def:tfMajLoss}. From the observation that $J_M^T < J_M^F$, we see that improved majority loss follows from an increased number of correctly classified majority datapoints; i.e., $> T_n \implies < J(S_M)$. While additional conditions are required to show the reverse is also true, improved majority loss following from improved classification is enough to support the approach of using improved majority loss to incentivize improved majority classification. The observation that $J_M^T < J_M^F$ follows from Lemma \ref{MAJlossInequality} below.

To begin, we recall from Definitions \ref{def:softmax} and \ref{def:loss} the connection between a majority datapoint's loss $l(s;z(w))$ and whether or not its predicted label $\hat{y}$ is correct or incorrect. With the following lemma, we extend the connection between loss and classification to show that loss of a correctly classified majority datapoint is strictly less than the loss of an incorrectly classified majority datapoint. 

\newtheorem{lemma}{Lemma}

\begin{lemma}
\label{MAJlossInequality}
Given a correctly classified majority datapoint $s^T \in S_M$, an incorrectly classified majority datapoint $s^F \in S_M$ and a neural network output $z(w) \in \mathbb{R}$, it follows that $l(s^T;z(w))< l(s^F;z(w)).$
\begin{proof}
Recall that with regards to classification, a majority datapoint $s \in S_M$ is classified either correctly as $\hat{y}=0$ or incorrectly as $\hat{y}=1$ as follows

\begin{align*}
    \hat{y} &= \begin{cases} 
        0 \iff \frac{e^{1-z(w)}}{ e^{z(w)} + e^{1-z(w)}} \geq \frac{e^{z(w)}}{ e^{z(w)} + e^{1-z(w)}}\\ 
        1 \iff \frac{e^{1-z(w)}}{ e^{z(w)} + e^{1-z(w)}} < \frac{e^{z(w)}}{ e^{z(w)} + e^{1-z(w)}}.
    \end{cases}\\
\end{align*}

Therefore since 
$\frac{e^{z(w)}}{ e^{z(w)} + e^{1-z(w)}},  \frac{e^{1-z(w)}}{ e^{z(w)} + e^{1-z(w)}} \in \mathbb{R}$ and $\frac{e^{z(w)}}{ e^{z(w)} + e^{1-z(w)}} + \frac{e^{1-z(w)}}{ e^{z(w)} + e^{1-z(w)}} = \frac{e^{z(w)} + e^{1-z(w)}}{ e^{z(w)} + e^{1-z(w)}} = 1$,

\begin{align*}
    \hat{y} &= \begin{cases} 
        0 \iff \frac{e^{1-z(w)}}{ e^{z(w)} + e^{1-z(w)}} \geq \frac{1}{2}\\ 
        1 \iff \frac{e^{1-z(w)}}{ e^{z(w)} + e^{1-z(w)}} < \frac{1}{2}
    \end{cases}\\
\end{align*}

from which it follows by definition since $\forall s \in S_M, l(s,z(w)) = -\log(\frac{e^{1-z(w)}}{ e^{z(w)} + e^{1-z(w)}})$

\begin{align*}
    \hat{y} &= \begin{cases} 
        0 \iff \frac{1}{e^{l((s^T;z(w))}} = \frac{e^{1-z(w)}}{ e^{z(w)} + e^{1-z(w)}} \geq \frac{1}{2}\\ 
        1 \iff \frac{1}{e^{l((s^F;z(w))}} = \frac{e^{1-z(w)}}{ e^{z(w)} + e^{1-z(w)}} < \frac{1}{2}
    \end{cases}\\
\end{align*}

and it concludes since 
$\frac{1}{e^{l((s^T;z(w))}} \geq \frac{1}{2} > \frac{1}{e^{l((s^F;z(w))}}$ with

\begin{align*}
    \frac{1}{e^{l((s^T;z(w))}} > \frac{1}{e^{l((s^F;z(w))}} \iff l(s^T;z(w))< l(s^F;z(w)).\\
\end{align*}
\end{proof}

By Lemma ~\ref{MAJlossInequality}, $J_M^T<J_M^F$. 
\end{lemma}

\section{Evaluation}
\label{evaluation}

\begin{small}
\begin{table*}[htb]
\centering 
    \begin{tabular}{c c c c c c c c c c}
     \toprule
     \textup{Dataset} & \text{Metric} & \text{w/o} & \text{SMOTE} & \text{B-SMO} & \text{S-SMO} & \text{SMO-GAN}  & \text{GBO} & \text{SSG} & \text{MUBO} \\
     \midrule
     Ionosphere & \textbf{F1} & 0.91 & 0.94 & 0.89 & 0.91 & \textbf{0.98} & 0.92 & 0.92 & 0.87 \\
      & \text{Precision} & 0.93 & 0.95 & 0.90 & 0.91 & 1.00 & 0.94 & 0.94 & 0.91 \\
      & \text{Recall} & 0.89 & 0.93 & 0.88 & 0.91 & 1.00 & 0.91 & 0.91 & 0.85 \\
     \midrule
     Gisette & \textbf{F1} & 0.89 & 0.88 & 0.85 & 0.94 & - & 0.87 & 0.83 & \textbf{0.96}\\
      & \text{Precision} & 0.97 & 0.94 & 0.98 & 0.96 & - & 0.97 & 0.96 & 0.95 \\
      & \text{Recall} & 0.84 & 0.84 & 0.78 & 0.93 & - & 0.81 & 0.76 & 0.97 \\
     \midrule
     Abalone & \textbf{F1} & 0.84 & 0.76 & 0.75 & 0.82 & 0.76 & 0.83 & 0.83 & \textbf{0.86} \\
      & \text{Precision} & 0.85 & 0.74 & 0.74 & 0.79 & 0.80 & 0.80 & 0.80 & 0.86 \\
      & \text{Recall} & 0.83 & 0.84 & 0.85 & 0.86 & 0.75 & 0.88 & 0.88 & 0.86 \\
     \midrule
     Spambase & \textbf{F1} & 0.92 & 0.91 & 0.90 & 0.91 & 0.92 & \textbf{0.93} & \textbf{0.93} & 0.92 \\
      & \text{Precision} & 0.92 & 0.91 & 0.90 & 0.90 & 0.94 & 0.93 & 0.93 & 0.89 \\
      & \text{Recall} & 0.92 & 0.92 & 0.91 & 0.91 & 0.91 & 0.93 & 0.93 & 0.95 \\
     \midrule
     Connect4 & \textbf{F1} & 0.82 & 0.91 & 0.91 & 0.90 & 0.94 & 0.94 & 0.95 & \textbf{0.96} \\
      & \text{Precision} & 0.98 & 0.85 & 0.86 & 0.85 & 0.92 & 0.91 & 0.91 & 0.92 \\
      & \text{Recall} & 0.75 & 0.99 & 0.99 & 0.99 & 1.00 & 0.99 & 0.99 & 0.99 \\
     \bottomrule
    \end{tabular}
    \caption{Average $F1$, precision and recall scores provide a comparison of MUBO to a technique without sampling (w/o); three classic sampling techniques - SMOTE, Borderline-SMOTE (B-SMO) and SVM-SMOTE (S-SMO); and three state-of-the-art methods - SMOTified-GAN (SMO-GAN), GBO and SSG. Experiments across all methods are with $20\%$ testing data.}
    \label{fig:resultsTake2}
\end{table*}
\end{small}

In what follows, we present an evaluation of our proposed method that further supports our undersampling approach using bilevel optimization. As shown in Table \ref{fig:resultsTake2}, our method achieves performance results competitive with state-of-the-art sampling methods SMOTified-GAN, GAN-based Oversampling (GBO) and SVM-SMOTE-GAN (SSG). Additional analysis comparing minority versus majority classification results across MUBO, GBO and SSG shows that our method is stronger at classifying minority data in particular. 

\subsection{Experimental Set-up}
We evaluate our proposed method with five datasets from the UCI Machine Learning Repository (Kelly et al. 2023) ranging in size from 351 to 376,640 instances; in dimension from 8 to 5,000 features; and in imbalance ratio from 0.294\% to 39.39\% (Table \ref{fig:dataSets}). These benchmark datasets are commonly used for method development. For example, the state-of-the-art techniques to which we compare our approach - SMOTified-GAN, GBO and SSG - were also initially evaluated using four of the five datasets. The fifth dataset - Gisette - is of significantly higher dimension with 5,000 features.

MUBO's neural network has two hidden linear layers with 256 and 128 neurons per layer; a rectified linear unit function; an additional linear layer with 128 neurons; and a softmax layer for rescaling the output elements so that they lie between 0 and 1 and sum to 1. A binary cross entropy loss function is used to train the neural network with a data batch size of 32 and an initial learning rate of 0.0001 with the Adam optimizer. This set-up aligns with those found in the SMOTified-GAN, GBO and SSG methods \cite{sharmaSmotifiedGan2022, ahsanEnhancing2024}. To generate robust results, there were five runs for each dataset with a random seed generated at the beginning of each run. 

Developed in Python with PyTorch, code for seven methods - without sampling, SMOTE, Borderline-SMOTE, SVM-SMOTE, GBO, SSG and MUBO - was successfully run on a Linux-based cluster that uses an Intel(R) Xeon(R) CPU E5-2623 v4 2.60GHz processor and a NVIDIA GeForce GTX 1080 GPU card. Unable to execute SMOTified-GAN's experiments on our machine, we used their published results in Table \ref{fig:resultsTake2}. Our code is publicly available at \url{https://github.com/kkmedlin/MUBO}.
\subsection{Metrics}
In addition to the F1 scores previously defined in Definition \ref{def:F1}, the additional two metrics are used: 

\begin{align*}
    \text{Precision} &= \frac{1}{2}\left(\underbrace{\frac{T_p}{T_p + F_p}}_{\text{Prec}_m} + \underbrace{\frac{T_n}{T_n + F_n}}_{\text{Prec}_M}\right)\\
\end{align*}
\begin{align*}
    \text{Recall} &= \frac{1}{2}\left(\underbrace{\frac{T_p}{T_p + F_n}}_{\text{Rec}_m} + \underbrace{\frac{T_n}{T_n + F_p}}_{\text{Rec}_M}\right) \\
\end{align*}  
$\text{Prec}_m$ and $\text{Prec}_M$ denote minority and majority precision, and $\text{Rec}_m$ and $\text{Rec}_M$ denote minority and majority recall.

Both precision and recall are important for evaluating imbalanced data classification. To differentiate between precision and recall, imagine two bins - one bin for minority data and one bin for majority. Low precision indicates that data has been put into the wrong bin. Low recall, on the other hand, indicates that a bin is missing some of its own datapoints. For example, high precision is important for fraud detection so as to avoid the false accusation of fraud. High recall, on the other hand, is important for medical image classification as missing a diagnosis can have fatal consequences. As the harmonic mean between precision and recall, the F1 score captures both precision and recall and is the preferred metric for evaluating imbalanced data classification. \cite{brownlee2021machine}

\subsection{Numerical results and analysis}
\label{results}
\begin{small}
\begin{table*}[htbp]
\centering 
    \begin{tabular}{c c c c c}
     \toprule
     \textup{Dataset} & \text{Instances} & \text{Features} & \text{Minority class (\%)} & \text{Description} \\
     \midrule
     Ionosphere & 351 & 34 & 35.71 & Classification of radar returns from the ionosphere \\
     Gisette & 2,750 & 5,000 & 10 & 4s and 9s from MNIST dataset with added noise \\
     Abalone & 4,177 & 8  & 20.1 & Predict abalone age with physical measurements \\
     Spambase & 4,601 & 57 & 39.39 & Predict spam or non-spam \\
     Connect4 & 376,640 & 42 & 3.84 & Connect-4 positions \\
     \bottomrule
    \end{tabular}
    \caption{Description of datasets}
    \label{fig:dataSets}
\end{table*}
\end{small}

\begin{small}
\begin{table*}[htb]
\centering 
    \begin{tabular}{c c c c c c c c c c c}
     \toprule
     \textup{Dataset} & \text{Method} & F1 & \text{$F1_m$} & \text{$F1_M$} & \text{$Prec.$} & \text{$Prec_m$} & \text{$Prec_M$} & \text{$Recall$} & \text{$Rec_m$} & \text{$Rec_M$} \\
     \midrule
     Ionosphere & GBO & \textbf{0.92} & 0.90 & 0.95 & 0.94 & 0.96 & 0.92 & 0.91 & 0.85 & 0.98\\
      & MUBO & 0.87 & 0.81 & 0.93 & 0.91 & 0.92 & 0.89 & 0.85 & 0.74 & 0.97\\
     \midrule
     Gisette & GBO & 0.87 & 0.76 & 0.98 & 0.97 & 0.97 & 0.97 & 0.81 & 0.62 & 1.00\\
      & MUBO & \textbf{0.96} & 0.95 & 0.96 & 0.95 & 0.92 & 0.98 & 0.97 & 0.99 & 0.94\\
     \midrule
     Abalone & GBO & 0.83 &0.73 & 0.92 & 0.80 & 0.63 & 0.97 & 0.88 & 0.88 & 0.87\\
      & MUBO & \textbf{0.86} & 0.87& 0.86 & 0.86 & 0.84 & 0.88 & 0.86 & 0.89 & 0.83\\
     \midrule
     Spambase & GBO & 0.93 & 0.91 & 0.94 & 0.93 & 0.92 & 0.94 & 0.93 & 0.91 & 0.94\\
      & MUBO & \textbf{0.92} & 0.92 & 0.92 & 0.92 & 0.89 & 0.96 & 0.92 & 0.96 & 0.88\\
     \midrule
     Connect4 & GBO & 0.94 & 0.89 & 1.00 & 0.91 & 0.82 & 1.00 & 0.99 & 0.98 & 0.99\\
      & MUBO & \textbf{0.96} & 0.91 & 1.00 & 0.92 & 0.85 & 1.00 & 0.99 & 0.99 & 0.99\\
     \bottomrule
    \end{tabular}
    \caption{A comparison between minority classification versus majority classification, $F1$ (average), $F1_m$ (minority), $F1_M$ (majority), \text{$Prec.$} (average precision), \text{$Prec_m$} (minority), \text{$Prec_M$} (majority), \text{$Recall$} (average recall), \text{$Rec_m$} (minority) and \text{$Rec_M$} (majority) scores are provided for MUBO and the state-of-the-art method GBO. }
    \label{fig:minMAJresults}
\end{table*}
\end{small}
As shown in Table \ref{fig:resultsTake2}, the average F1 scores for our method surpass the average F1 scores of all classic and state-of-the-art sampling methods on the datasets Gisette, Abalone and Connect4. MUBO's leading F1 scores range from $0.86$ to $0.96$ (with variances from $1.31e{-5}$ to $2.29e{-5}$), and the next highest scores are $1\%$ to $2\%$ lower. The two datasets for which our approach does not lead -- Spambase and Ionosphere -- are, in fact, the least imbalanced. In comparison with state-of-the-art methods, MUBO met SMOTified-GAN's F1 score with the Spambase dataset while achieving an F1 score just shy of that of GBO and SSG at only $1\%$ lower. On the Ionosphere dataset, the average F1 score of our method is $11\%$ lower than that of SMOTified-GAN and $5\%$ lower than GBO. A deeper investigation into these results as shown in Table \ref{fig:minMAJresults} reveal that Ionosphere's majority F1 score is $0.93$ while its minority F1 score is $0.81$, indicating that our technique struggled to classify minority data. Ionosphere is the smallest dataset in the study with 351 instances. Given that our method utilizes all available minority data, Ionosphere's experimental results indicate that a small dataset size is a hindrance to our undersampling approach.

Table \ref{fig:minMAJresults} provides a comparison between minority and majority classification performance. On all five datasets, $F1_m < F1_M$ for GBO, signaling that this minority oversampling method reliant on synthetic data is better at classifying majority data than minority data. In contrast, our technique does not use synthetic data and, therefore, sidesteps potential hurdles to minority data classification. Aside from Ionosphere, our method's minority F1 score surpasses the minority F1 score for GBO, with the next highest score $1\%$ to $19\%$ lower. Our approach exhibits less optimal performance on majority recall. Given that the majority subset with which MUBO trains is significantly smaller than the full set of majority data with which GBO trains, a lower majority recall score makes sense.

Lastly, our method performs similarly with respect to minority and majority data; i.e., $F1_m \sim F1_M$. With the upper level optimization problem focused solely on finding an optimal subset of majority training data, our technique re-tilts its classification abilities back to center. Recalling that $F1 = \frac{1}{2}(F1_m + F1_M)$, we see that both high $F1_m$ and high $F1_M$ scores are required to achieve average F1 scores that dominate the rankings. By eliminating hindrances to minority data classification and providing support for both minority and majority data classification, our method improves the classification of imbalanced data.

\section{Conclusion and Future Outlook}
Both oversampling and undersampling methods aim to attain balanced training sets; however, problems of noise and overlap  persist in oversampling methods that use synthetic data. Our proposed method's strong results on larger datasets can be attributed to its approach that (i) undersamples majority data instead of oversampling minority data and (ii) uses a bilevel optimization framework to attain an optimal subset of majority training data. 

Additional conditions are required to formally prove experimental results showing improved majority loss implies improved average F1 score. Future work includes the formulation of subsequent methods that provide conditions for proving improved results. To add, the formulation of subsequent methods also includes a generalization of current work to include strong results on both small and large datasets.

\section{Acknowledgments}
This material is based upon work supported by the U.S. Department of Energy, Office of Science, Office of Workforce Development for Teachers and Scientists, Office of Science Graduate Student Research program under contract number DE-SC0014664. This work was also supported in part by the U.S. Department of Energy, Office of Science, Office of Advanced Scientific Computing Research, Scientific Discovery through Advanced Computing (SciDAC) Program through the FASTMath Institute, RAPIDS institute, distributed resilience systems for science program and the integrated computational and data infrastructure for science discovery program under Contract No. DE-AC02-06CH11357.

\bibliographystyle{unsrt}

\end{document}